%% file: arxiv.tex
\documentclass{article}

\PassOptionsToPackage{numbers, compress}{natbib}
\usepackage[preprint]{neurips_2026}

\usepackage[utf8]{inputenc} %
\usepackage[T1]{fontenc}    %
\usepackage[colorlinks=true, allcolors=RoyalBlue]{hyperref}       %
\usepackage{url}            %
\usepackage{booktabs}       %
\usepackage{amsfonts}       %
\usepackage{nicefrac}       %
\usepackage{microtype}      %
\usepackage{xcolor}         %

\usepackage{amsthm}
\usepackage{comment}
\usepackage[color=white,bordercolor=white,backgroundcolor=white,linecolor=white]{todonotes}
\usepackage{enumitem}
\usepackage[most]{tcolorbox}
\usepackage{titletoc}        %

\usepackage{tikz}
\usetikzlibrary{shapes.geometric, arrows.meta, positioning, fit, backgrounds, calc}
\usepackage{rotating}
\usepackage{gunnarmacros}

\definecolor{treatmentcolor}{HTML}{D42A7E}
\definecolor{outcomecolor}{HTML}{3E9C74}
\definecolor{recipecolor}{HTML}{6161D2}
\definecolor{effectcolor}{HTML}{e6ab02}
\definecolor{controlcolor}{HTML}{e6ab02}

\newcommand{\treatment}{\textcolor{treatmentcolor}{a}}
\newcommand{\outcome}{\textcolor{outcomecolor}{y}}
\newcommand{\recipe}{\textcolor{recipecolor}{\psi}}
\newcommand{\effect}{\tau}
\newcommand{\control}{\textcolor{controlcolor}{\emptyset}}

\title{Validity Threats for Foundation Model Research}

\author{
  Gunnar König\textsuperscript{1} \quad
  Martin Pawelczyk\textsuperscript{2} \quad
  Ulrike von Luxburg\textsuperscript{1} \quad
  Sebastian Bordt\textsuperscript{1} \\[0.6em]
  \textsuperscript{1}University of Tübingen, Tübingen AI Center \quad
  \textsuperscript{2}University of Vienna
}
\begin{document}

\maketitle

\begin{abstract}

\input{neurips/sections/abstract}

\end{abstract}

\input{neurips/sections/introduction}

\input{neurips/sections/setup}

\input{neurips/sections/proxy}

\input{neurips/sections/observational}

\input{neurips/sections/singlerun-v5}

\input{neurips/sections/validity-profiles}

\input{neurips/sections/related_work}

\section{Discussion}

In this work, we provide an evaluation framework for validity threats for foundation model research. In practice, researchers may often want to know how to address particular validity threats in their research designs---can clever experiments, better theoretical controls, or additional data help address the validity threats? One answer to this question is that we can always perform the ideal experiment, and see if results agree (for example, by training a large model we can verify the external validity of a scaling law across scale). There are other answers as well, for example, a scientific literature can over time accumulate knowledge about the practical importance of different threats to different research questions and designs (e.g., we know that scaling laws for the cross-entropy loss after pretraining work pretty well, but that this does not necessarily hold for all other outcomes). It could also be the case that we have theory which supports the validity of the experimental design \citep{yang2021tuning,shulgin2026deriving}. Open-science initiatives like the \href{https://marin.community/}{Marin Project} that openly document training recipes and meta-data can also help. At any rate, however, the necessary first step is always to be {\it aware} of potential validity threats, and to have a common language to discuss them---and this is exactly what we provide with our framework.

\section*{Acknowledgments}

This work has been supported by the German Research Foundation through the Cluster of Excellence ``Machine Learning -- New Perspectives for Science'' (EXC 2064/1 number 390727645).

\bibliographystyle{plainnat}
\bibliography{neurips/references}

\appendix

\input{neurips/sections/supplement}

\end{document}

%% file: neurips/sections/abstract.tex
Controlled experiments are the backbone of machine learning research, but at the scale of modern foundation models, they have become prohibitively expensive.
Instead, the community increasingly relies on research strategies that approximate the ideal experiment at a fraction of the cost: proxy experiments and scaling laws, observational studies with publicly available models, and single-run designs that leverage variation within individual training runs.
In this work, we argue that there is no free lunch when approximating large-scale experiments on a compute budget. Specifically, savings in compute come at the cost of \emph{validity threats}---hidden and sometimes untestable assumptions that, when violated, can invalidate research claims. To help navigate such threats, we propose an evaluation framework that casts foundation model research as a causal inference problem. Within this framework, we evaluate different research strategies through four types of validity adapted from the empirical social sciences---statistical, internal, external, and construct validity. 
We find that each strategy comes with a characteristic {\it validity profile}: proxy experiments trade external and construct validity for statistical and internal validity; observational studies face confounding and effect heterogeneity; and single-run designs are strained by interference between treated units.
This analysis reveals several validity threats that have received insufficient attention in the literature.  
Overall, our evaluation framework provides researchers with a practical toolkit for scrutinizing validity threats in foundation model research~designs.

%% file: neurips/sections/introduction.tex
\section{Introduction}

With the advent of foundation models, the computational cost of training state-of-the-art machine learning models has increased dramatically \citep{bommasani2021opportunities,cottier2024rising}. In particular, training a frontier language model now requires tens of thousands of GPUs \citep{brown2020language,dubey2024llama,sevilla2024scaling}. %
This trend has profound implications for both model developers, who need to make design decisions at a scale so large that models can only be trained once, and for researchers, who aim to answer research questions about models that they cannot afford to train themselves. 
To work around this cost, the community has developed various strategies to study model behavior on a smaller compute budget \citep{pawelczyk2024machine,shi2024muse,maini2024tofu,dorna2025openunlearning,lesci2025causal,bordt2026train}.
For example, developers use scaling laws to predict optimal hyperparameters from small-scale experiments \citep{kaplan2020scaling,hoffmann2022training,tao2024scaling,bergsma2025power,shukor2025scaling},
or analyze public meta-data to determine factors that matter for downstream performance \citep{thrush2024improving,liu2025not}. 

In this work, we argue %
that there is no free lunch when approximating foundation model experiments on a smaller compute budget, because the \textbf{savings in compute come at the cost of hidden and sometimes untestable assumptions}. For example, scaling laws must rely on the assumption that the results generalize to the target scale, and observational methods must assume that there is no confounding. If these assumptions are violated, then research results can become invalid, rendering them fundamentally incorrect or inapplicable. In other words, the assumptions present {\it threats to the validity} of the research. While some works discuss validity threats in great detail \citep{thrush2024improving,zhang2025benchmark}, this discussion is also absent from many papers.
We hypothesize that this is because the methodological and causal challenges---though familiar in other disciplines---are relatively new to empirical machine learning research \citep{wallach2024evaluating}.

\begin{figure}[t]
  \centering
  \includegraphics[width=0.99\linewidth]{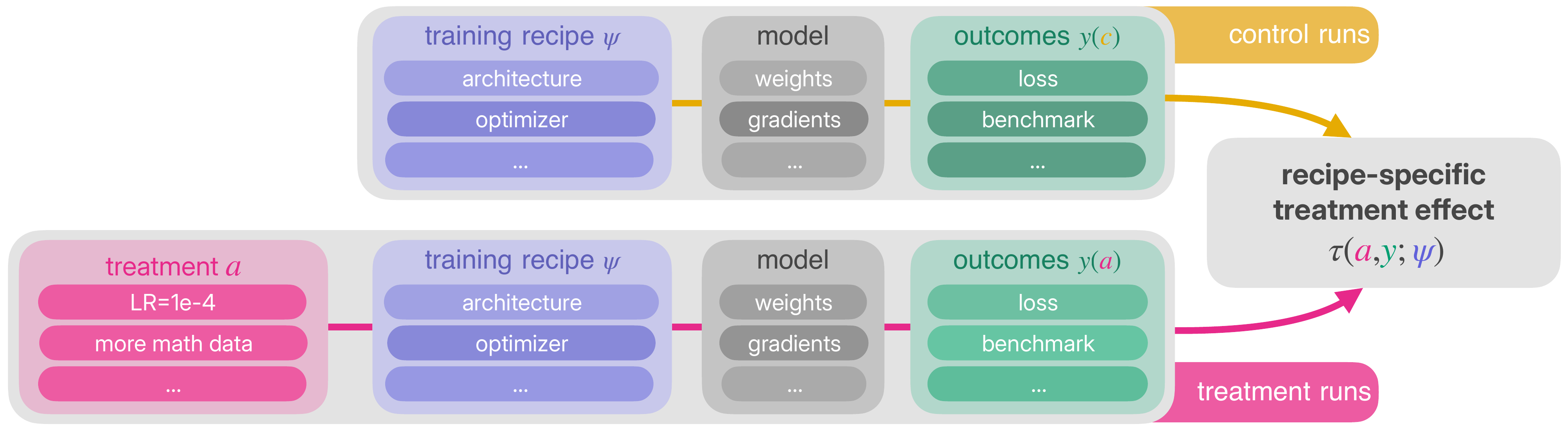}
  \caption{{\bf Foundation model research as a causal inference problem.}
  To understand and improve foundation models, we need to answer questions about how
\textcolor{treatmentcolor}{changes} to a \textcolor{recipecolor}{training recipe} affect \textcolor{outcomecolor}{outcomes} of interest---for example, how \textcolor{treatmentcolor}{adding math data} affects \textcolor{outcomecolor}{reasoning benchmark scores}. 
At frontier scale, the ideal experiment that identifies this \textbf{treatment effect} must be approximated.
  }
  \label{fig:overview}
\end{figure}

\textbf{An evaluation framework for foundation model research.} 
To address the challenge of validity threats in foundation model research, we propose an evaluation framework for different research designs  (Section~\ref{sec:the_framework}). At a high level, our framework formulates foundation model research as a causal inference problem:
the goal is to approximate ideal but infeasible experiments that measure how changes to the training recipe affect outcomes of interest (Figure \ref{fig:overview}).
For example, finding the optimal learning rate for a large-scale training run is formalized as estimating the effects of treatments that set the learning rate to different values. To help identify validity threats in any given design, our framework decomposes the analysis into two layers: Research strategies and validity types.

{\bf Research strategies.} We identify a set of general research strategies that often serve as building blocks in the research designs of individual works: The {\it proxy approach} (Section \ref{sec:proxys}) encompasses research where experiments are conducted on cheaper proxies---smaller models, cheaper treatments, or cheaper outcomes---and findings are extrapolated to the target setting. The {\it observational approach} (Section \ref{sec:observational}), in analogy to observational studies in other fields, describes approaches that leverage the ecosystem of publicly available models to study the relationship between training choices and outcomes without conducting new training runs. In the {\it single run approach} (Section \ref{sec:single-run}), researchers extract insights by exploiting the internal structure of a single training run---for example, by treating different training data points as independent units that can be used to form treatment and control populations. While it may not provide a fully exhaustive categorization, the majority of recent works utilize one or more of these three research strategies.

{\bf Validity types.} Building on a framework for causal inference from the social sciences \citep{shadish2001validity}, we examine different research strategies through the lens of four types of validity: \emph{statistical validity} asks whether an estimate is reliable given the finite sample; \emph{internal validity}, whether it reflects a genuine causal effect; \emph{external validity}, whether findings generalize to the target training recipe; and \emph{construct validity}, whether the operationalization of treatment and outcome reflects the intended research question.

{\bf Validity profiles.} Based on our discussion of the validity threats inherent to different research strategies, we propose a matrix of validity profiles that describe which types of validity are most at risk with different strategies (Section \ref{sec:validity_profiles}). In these profiles, we classify validity types as being rather not affected (\bgreen), requiring careful consideration (\byellow), or key threat (\bred). While our framework and the four validity types provide a general vocabulary for the different ways in which a research design can fail, the proposed validity profiles offer concrete guidance for evaluation. Taken together, this provides researchers with a {\bf practical toolkit} for evaluating validity threats for foundation model research.

%% file: neurips/sections/setup.tex
\section{An Evaluation Framework for Foundation Model Research} 
\label{sec:the_framework}
In this section, we introduce our framework.
We formalize empirical foundation model research as a causal inference problem 
and introduce four types of validity that any research design must establish.

{\bf The training recipe and outcome measures.} A foundation model is trained according to a particular \textcolor{recipecolor}{\textbf{training recipe} $\recipe$}, which encompasses all controllable parts of model training (the architecture, the training data, all random seeds, and so on). 
Up to some uncontrollable random variation, the training recipe determines the \textbf{foundation model} along with a range of \textcolor{outcomecolor}{\textbf{outcome measures} $\outcome$} (including upstream outcomes like the training loss, and downstream outcomes such as benchmark scores).
When models share central parts of their recipe, we often say they belong to the same \textbf{family}.

{\bf Foundation model research as a causal inference problem.}
The goal of foundation model research is to develop a training recipe that  yields high-quality outcomes. 
To formally study this problem, suppose that there exists a fixed \textcolor{recipecolor}{training recipe $\recipe$} that we want to improve. 
Now, we need to answer questions about how \textcolor{treatmentcolor}{changes in the training recipe} would affect the \textcolor{outcomecolor}{outcomes}, for example:
\begin{itemize}[leftmargin=0.60cm, itemsep=1pt, topsep=1pt, parsep=0pt]
    \item[1.] Would \textcolor{treatmentcolor}{\underline{\textcolor{black}{decreasing the learning rate}}} improve \textcolor{outcomecolor}{\underline{\textcolor{black}{downstream performance}}}?
    \item[2.] Would \textcolor{treatmentcolor}{\underline{\textcolor{black}{adding a data source during mid-training}}} improve \textcolor{outcomecolor}{\underline{\textcolor{black}{reasoning behavior}}}?
    \item[3.] Would \textcolor{treatmentcolor}{\underline{\textcolor{black}{increasing the size of the model}}} improve \textcolor{outcomecolor}{\underline{\textcolor{black}{post-train-ability}}}?
\end{itemize} %

These are {\it causal} questions: 
We aim to quantify the effect of a \textcolor{treatmentcolor}{\textbf{treatment~$\treatment$}}  on the \textcolor{outcomecolor}{outcomes~$\outcome$} for a particular \textcolor{recipecolor}{recipe~$\psi$}. 
In Definition~\ref{def:ate-psi} we make this precise.
To do so, we adopt the potential outcomes framework \citep{rubin1974estimating} and write $\outcome(\treatment)$ for the treatment outcome and $\outcome(\control)$ for the control outcome, that is, the outcome for the unmodified recipe $\recipe$.
We visualize Definition~\ref{def:ate-psi} in Figure~\ref{fig:overview}.

\begin{definition}[\bf Recipe-specific treatment effect]
\label{def:ate-psi}
Let \textcolor{recipecolor}{$\recipe$} be a training recipe, $\treatment$ a treatment, and $\textcolor{outcomecolor}{\outcome}(\textcolor{treatmentcolor}{\treatment})$ and $\textcolor{outcomecolor}{\outcome}(\control)$ the treatment and control outcomes.
Then the recipe-specific treatment effect is defined as:
\begin{equation*}
  \effect(\treatment, \outcome; \recipe) = \E{\outcome(\treatment) \mid \recipe} - \E{\outcome(\control) \mid \recipe}.
\end{equation*}
\end{definition}
At frontier model scale, the ideal experiment that identifies Def.~\ref{def:ate-psi} is computationally prohibitive---every practical research design must approximate it, introducing assumptions whose violation can invalidate the results.
The goal of our framework is to help the reader systematically evaluate the central \emph{validity threats} for a particular design.
To this end, we build on a classical taxonomy from causal inference in the social sciences \citep{shadish2001validity} that distinguishes four types of validity:
\begin{enumerate}[leftmargin=0.60cm, itemsep=1pt, topsep=1pt, parsep=0pt]
    \item \textbf{Statistical Validity}:
    Each approach must estimate some parameter $\beta$ from a finite, noisy sample.
    Statistical validity asks whether the estimate $\hat{\beta}$ reliably reflects the true $\beta$. 
	\item \textbf{Internal Validity}:
    Granted a reliable estimate $\hat{\beta}$, does it reflect a genuine \emph{causal} effect $\tau$?
    In well-designed experiments, association and causation coincide, however, this is not the case in observational settings.
    For example, when recent models with a new architecture achieve better outcomes, that does not mean that the architecture caused the improvement.%
    \item \textbf{External Validity}:
    A genuine causal effect $\tau$ is only informative if it applies to the \textcolor{recipecolor}{recipe $\recipe$} that we care about. 
    External validity is about the gap between $\tau(\, \cdot \,; \recipe)$ and $\tau( \, \cdot \,; \recipe')$---for example, between a treatment effect for a small model versus the treatment effect at scale.
	\item \textbf{Construct Validity}: 
    A genuine causal effect $\tau$ is only informative if the operationalization of \textcolor{treatmentcolor}{treatment $\treatment$} and \textcolor{outcomecolor}{outcome $\outcome$} actually reflects the constructs in the research question.
    For example, changing the treatment from ``pre-training on private data'' $\treatment$ to ``fine-tuning on private data'' $\treatment'$ may threaten construct validity.
\end{enumerate}

Over the course of the paper, we discuss the validity threats introduced by the different research strategies at an intuitive level, accompanied by a formal discussion of the assumptions made by different strategies in Appendix~\ref{app:formal}.
A glossary of important terms can be found in Appendix~\ref{app:glossary}.

%% file: neurips/sections/proxy.tex
\section{The Proxy Approach}
\label{sec:proxys}

\begin{figure}[t]
    \centering
    \includegraphics[width=0.99\linewidth]{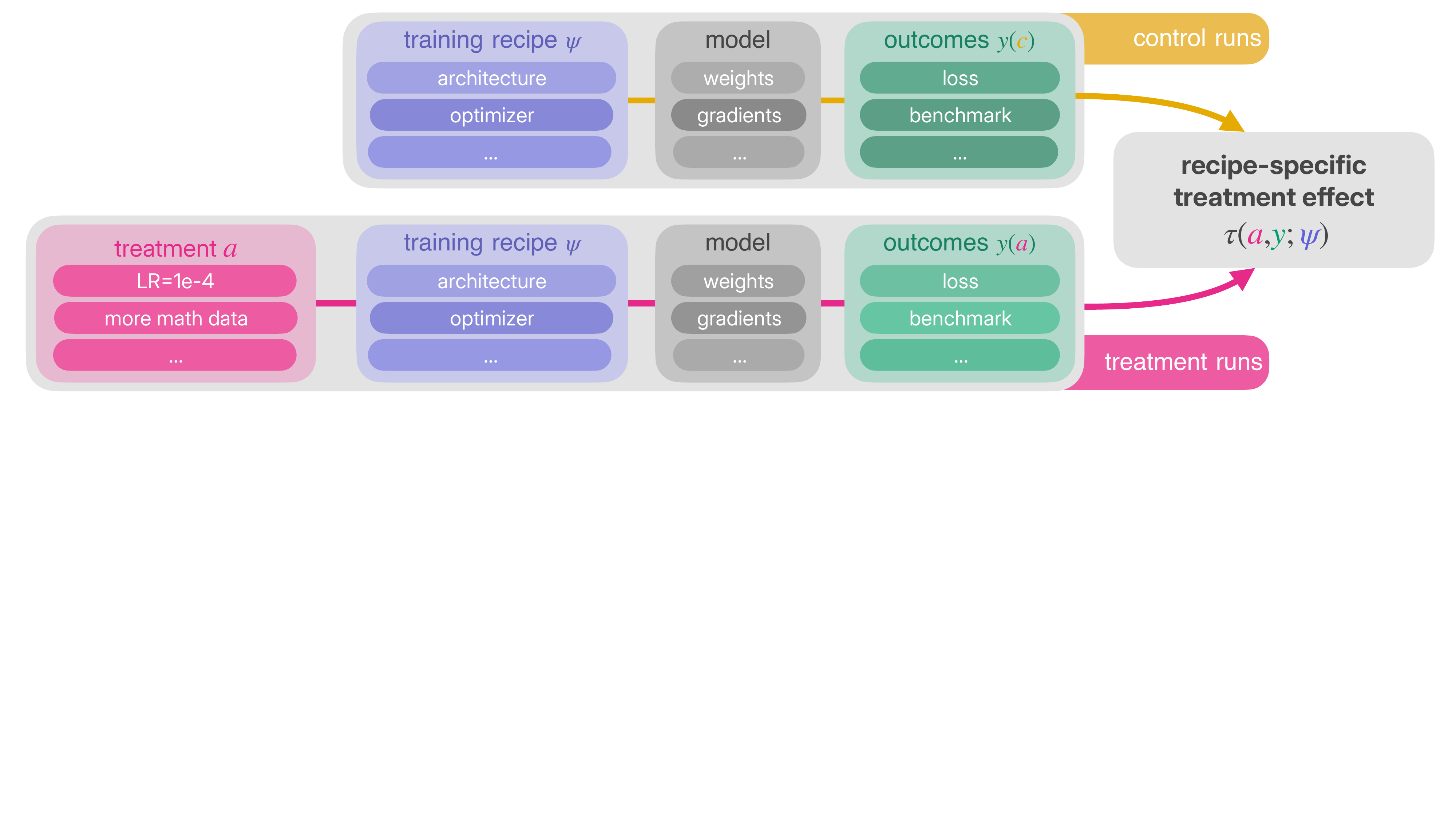}
    \caption{{\bf The proxy approach.} 
    We conduct an experiment but replace one or several parts with proxies (\dashedline) to make it computationally feasible.
    For example, we conduct the experiment with a \textcolor{recipecolor}{proxy model} (1B instead of 90B parameters), a \textcolor{treatmentcolor}{proxy treatment} (fine-tuning instead of pre-training on math data), or a \textcolor{outcomecolor}{proxy outcome} (validation loss instead of task performance).
    }
    \label{fig:proxy-approach}
\end{figure}

In this section, we begin our discussion of the validity threats of different research strategies with the proxy approach. In the proxy approach, a controlled experiment is performed, however some part of the experiment---the \textcolor{recipecolor}{model}, the \textcolor{treatmentcolor}{treatment}, or the \textcolor{outcomecolor}{outcome}---is replaced with a proxy to make it computationally feasible. This is illustrated in Figure \ref{fig:proxy-approach}: The left part of Figure \ref{fig:proxy-approach} illustrates the ideal experiment, whereas the right part illustrates various proxies. 


\input{neurips/sections/proxy_models}

\input{neurips/sections/proxy_treatments}

%% file: neurips/sections/proxy_models.tex
\subsection{Proxy Models}
\label{sec:proxy:models}

Perhaps the best known approach to overcome the problem that one cannot repeatedly train a foundation model is to train smaller models instead. We use the term {\it proxy model} to refer to models that are trained according to the original training recipe  but with (orders of magnitude) less compute. Proxy models usually follow a particular {\it scaling recipe} that dictates how  hyperparameters such as the learning rate are adjusted with scale. This means that there are  \textcolor{recipecolor}{proxy recipes $\recipe'$} that are scaled-down versions of the original \textcolor{recipecolor}{training recipe $\recipe$}.\footnote{The Glossary in Supplement \ref{app:glossary} gives an overview of central terms like training recipe and scaling recipe used throughout.}  %

A straightforward way to employ proxy models is under the assumption that the treatment effect does not depend on the considered scale \citep{magnusson2025datadecide,thudi2025mixmin}. 
A particular variant of this is {\bf under-training}, where the parameter count is kept the same, but the number of training tokens is reduced \citep{zhang2025persistent,bordt2026train,o2025deep}. The more interesting scenario, however, is when the treatment effect depends on the scale in a way that is predictable \citep{hestness2017deep}. For example, the optimal learning rate usually decreases with the scale of the model \citep{haas2025surprising,shulgin2026deriving}. If this is the case, then the goal is to estimate a {\it scaling law} that can accurately {\it predict optimal training recipes at a large-scale using models trained at a small scale.}

{\bf Scaling Laws.} \citet{kaplan2020scaling} and \citet{hoffmann2022training} observed that the validation loss after pretraining follows a power-law relationship in the sizes of model and training set, and that this relationship could be used to predict the optimal training recipe under a fixed compute budget. Subsequently, researchers have proposed scaling laws for many important components of the training recipe, including the batch size, learning rate, and training data composition \citep{bi2024deepseek,bergsma2025power,shukor2025scaling,tao2024scaling}. %

\textbf{Success and Failure of Proxy Models.}  Many works have demonstrated that proxy models can be used to predict outcomes \emph{across scale}. This is particularly true for predicting the pretraining validation loss as model and dataset size increase \citep{dey2023cerebras,hu2024minicpm,dubey2024llama}. For example, \citet[Figure 5]{bi2024deepseek} accurately predict the performance of their 67B parameter model using proxy models trained with 1000x less compute.
However, the literature also contains many examples where predictability breaks down, especially \emph{across different training recipes} \citep{muennighoff2023scaling,kim2025pre,mayilvahanan2025llms,longpre2025atlas}. For example, multiple works report that scaling laws for model and dataset size hold only as long as the composition of the pretraining data remains the same \citep{bi2024deepseek,goyal2024scaling,ye2024data}. Similarly, \citet{wang2025can}  show that the validity of proxy models for data selection breaks down once other training parameters change.

An important conceptual challenge for proxy experiments are tasks that exhibit {\it emergent behaviors} \citep{wei2022emergent,schaeffer2023emergent}. By definition, an emergent behavior cannot be studied with proxies below the threshold at which the behavior starts to emerge. In this respect, it is noteworthy that most scaling laws target the pretraining validation loss, which is known to scale smoothly in model and dataset size. However, a number of works provide scaling laws for other outcomes, in particular benchmark scores, sometimes successfully \citep{krajewski2025revisiting}, and sometimes with mixed results \citep{lourie2025scaling}.

{\bf Validity Threats.} {\bred}~The key validity threat for proxy models is whether it is valid to extrapolate the results  \emph{across scale} or even \emph{across training recipes}---a question of {\it external validity} (Assumption~\ref{ass:external-validity:proxy-recipe}). Importantly, predictability across scale and training recipes can {\it not} be determined based on experiments with proxy models alone---it has to be verified by conducting additional experiments at a larger scale. In this context, we would like to highlight that the statement that there is a ``scaling law'' is usually a {\it claim} for the external validity of proxy experiments across scale. 
{\bgreen}~When proxy models are combined with controlled experiments, estimates reflect a genuine causal effect (\emph{internal} validity). Similarly, \emph{statistical} validity can be obtained by training enough proxies. And because proxy models use the original treatment and outcome, they typically do not introduce \emph{construct} validity threats.

%% file: neurips/sections/proxy_treatments.tex
\subsection{Proxy Treatments and Outcomes}
\label{sec:proxy:treatments}

Instead of replacing the model with a proxy, we may also replace the \textcolor{treatmentcolor}{{treatment}} or \textcolor{outcomecolor}{outcome}. 
While this is conceptually similar to proxy models, proxy treatments and outcomes face different threats to their validity: Because we evaluate the target model at the target scale, the primary risk is no longer whether the findings generalize, but whether they are measuring the right phenomenon to begin with.
As we illustrate below, substituting the treatment or the outcome often fundamentally alters the causal estimand.

\textbf{Proxy treatments.}
Proxy treatments are common for research questions that involve adding or removing parts of the training data, where the ideal experiment would require to re-train the model from scratch.
To reduce cost, researchers approximate re-training with a proxy, for example by continuing to train an intermediate model checkpoint \citep{olmo20242}, or fine-tuning the fully trained model (see \citep{maini2024tofu,pawelczyk2024machine} among many examples).
As we illustrate with the case of fine-tuning proxies below, the replacement risks measuring the wrong causal effect.

{\bf Fine-tuning experiments overestimate privacy leakage.} 
Current privacy and unlearning research often relies on multi-epoch fine-tuning as a proxy for single-epoch pre-training to evaluate model behavior (e.g., \citep{pawelczyk2024machine,shi2024muse,maini2024tofu,dorna2025openunlearning}).
This proxy treatment has fundamentally different memorization dynamics:
Empirical and theoretical results suggest that training for more epochs increases privacy leakage \citep{song2021systematic,dong2022gaussian,leemann2023gaussian,tan2023blessing,zhang-etal-2024-order}.
In contrast, examples seen early in the one-epoch pretraining regime are typically forgotten quickly \citep{jagielski2023measuring,bordt2026train}.
Therefore, fine-tuning proxy treatments tend to overestimate the model's susceptibility to privacy attacks \citep{shokri2017membership,carlini2021extracting}.
When we rely on them, we measure a different causal effect, and risk accepting unnecessary performance trade-offs to achieve privacy.

\textbf{Proxy outcomes.} A parallel argument can be made for proxy outcomes:
Employing them risks answering the wrong causal question.
For example, benchmark prediction methods approximate otherwise costly benchmark scores by evaluating the model on a small subset of the benchmark data \citep{vivek2024anchor,li2024active,polo2024tinybenchmarks,pacchiardi2024100}.
Recent work shows that these approximations fail for the most performant models \citep{zhang2025benchmark}; optimizing against this distorted construct risks overlooking the best of all models.

\textbf{Validity Threats.} {\bred}~When proxy treatments/outcomes operationalize different constructs than their original counterparts, we risk measuring the \emph{wrong} causal effect. This is a threat to \emph{construct} validity (Assumptions~\ref{ass:construct-validity:proxy-treatment}--\ref{ass:construct-validity:proxy-outcome}). {\bgreen}~Proxy treatments and outcomes per se do not threaten \emph{statistical}, \emph{internal}, or \emph{external} validity.

%% file: neurips/sections/observational.tex
\section{The Observational Approach}
\label{sec:observational}

\begin{figure}[t]
\centering
\includegraphics[width=1.02\linewidth]{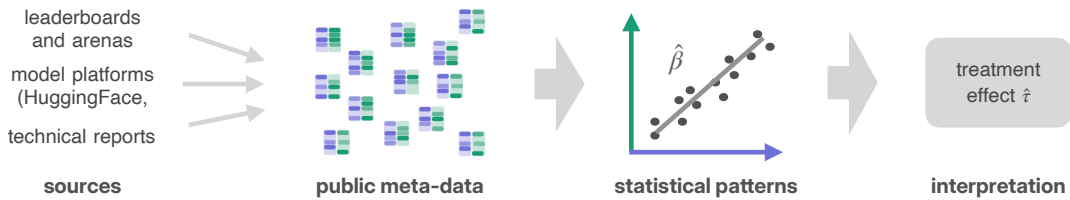}
\caption{{\bf The observational approach.} Instead of running new experiments, we can leverage the publicly available meta-data of existing training runs.
This data is observational, meaning that we cannot perform treatments and observe their effects.
Instead we have to use the statistical relationships between \textcolor{recipecolor}{recipe components} and \textcolor{outcomecolor}{outcomes} in combination with (untestable) causal assumptions to \emph{identify} treatment effects.
}
\label{fig:manyruns}
\end{figure}

In this section, we continue our discussion on the validity threats of different research strategies with the observational approach. In contrast to the proxy approaches discussed in Section \ref{sec:proxys}, the observational approach tries to avoid spending compute on model training. Instead, it analyzes the meta-data of existing models, thereby leveraging the millions of dollars collectively spent on their training. In recent works, the observational approach has been employed to derive scaling laws \citep{owen2024predictable,ruan2024observational,choshen2024hitchhiker,dominguez2024training,zhang2025train,lourie2025scaling,ho2024algorithmic,xiao2025densing}, to improve the pretraining data mix \citep{thrush2024improving}, and to predict benchmark performance from fewer samples \citep{li2024active,pacchiardi2024100,zhang2025benchmark}, to name only a few \citep{liu2025not,mertens2026there,li2026incompressibleknowledgeprobesestimating}. Formally, the observational approach operates by constructing datasets where each training run $j$ is one observation, the observed training recipe components are the ``features'' $\recipe^{(j)}$, and the outcomes are the ``labels'' $\outcome^{(j)}$.
The most common variant of the observational approach fits statistical models
to predict ``labels'' from ``features'' without employing explicit causal inference methodology (Figure~\ref{fig:manyruns}).
However, we would eventually like to arrive at valid causal inferences.

{\bf The Fragmented Meta-Data Landscape.}
Perhaps the most important consideration in the observational approach is the public availability of data about foundation models. For many model families meta-data are publicly available, in particular for open-weight models  \citep{biderman2023pythia,groeneveld2024olmo}. However, 
because there are few large-scale models, the sample size often remains small; The data is scattered across leaderboards and technical reports \citep{epoch2024notable,open-llm-leaderboard-v2,chiang2024chatbot,liu2025not};
And, importantly, %
much of the metadata remains undisclosed.
For proprietary models even the parameter counts are routinely kept secret \citep{li2026incompressibleknowledgeprobesestimating}---
but also for open-weight models details such as the pretraining data mix are commonly missing \citep{thrush2024improving,liu2025not,mertens2026there}.
To work around these limitations, the observational approach is often combined with proxies \citep{thrush2024improving,liu2025not,dominguez2024training}, such as per-domain-loss instead of the true mixing weights, inheriting the construct validity problems discussed in Section \ref{sec:proxy:treatments}.

{\bf Small Samples and Statistical Estimation.}
Since predicting outcomes from the full recipe is infeasible---e.g., the data mixing weights alone can comprise thousands of components, yet the number of fully documented training runs is small \citep{thrush2024improving,liu2025not}---%
researchers typically drop ``non-treatment features''
and impose simple parametric functional forms.
For example, \citet{ruan2024observational} assume sigmoidal relationships between scale parameters and downstream outcomes, and \citet{thrush2024improving} reduce their analysis to pairwise rank correlations.
When the treatment is low-dimensional and the parametric assumptions hold, the statistical validity of the observational approach \emph{can} compare favorably,
since it can draw on meta-data from multiple families.

{\bf Confounding and Exchangeability.}
But even when the statistical model is accurate, 
it only describes the \emph{associations} between features and outcomes.
Eventually, we would like to arrive at causal conclusions, but in observational data, association and causation do not necessarily coincide.
For example, so-called \emph{confounders} may cause both the treatment and the outcome variable, thereby inducing an association between them, even if the treatment has no causal effect.
A prominent confounder is \emph{calendar time} \citep{dominguez2024training,zhang2025train}:
Observational data shows that newer model families require much less compute to achieve a given benchmark performance \citep{dominguez2024training}.
This could be attributed to architectural or algorithmic innovations.
However, not only the architecture has evolved over time, but also the training data.
A range of evidence suggests that the data mix for newer models is increasingly tuned to reflect benchmark tasks \citep{dekoninck2024constat,dominguez2024training,zhang2024careful,zhang2025train}.
As a result, architectural developments are confounded with downstream outcomes via calendar time and training data composition (Figure \ref{fig:observational:graph}).

More formally, confounding is a violation of \emph{exchangeability} (Assumption~\ref{ass:exchangeability}), which requires that the potential outcomes are independent of the treatment assignment.
Exchangeability is \emph{untestable} based on observational data alone;
whether it holds can only be assessed using experiments or prior knowledge about the data-generating process. 
As examples in the literature illustrate \citep{dominguez2024training,singh2025leaderboard,whitfill2025note}, 
there is substantial reason to believe that this assumption is violated in practice---but nevertheless the issue was largely ignored.
In principle, \emph{causal inference} methodology allows to relax the exchangeability assumption:
Using prior knowledge about the data-generating process we can systematically remove non-causal associations \citep{rubin1974estimating,pearl2009causality}.
But so far, this methodology has found surprisingly little application. 

We suspect that in part data limitations are to blame: To adjust for a confounder, the confounder must be observed.
To work around these data limitations, a recent approach combines an observational sample with a proxy treatment:
To show that ``training on the test task'' is an important driver of improvements in scaling efficiency, 
\citet{dominguez2024training} fine-tune the models in their observational sample on task relevant data. 
While powerful, this strategy requires careful interpretation. The research question shifts from ``how does pretraining affect capability'' to ``how adaptable is the model to the test task'', a threat to \emph{construct validity} (Section \ref{sec:proxy:treatments}).

{\bf Effect Heterogeneity across Families.}
As discussed in Section~\ref{sec:proxy:models}, treatment effects can vary substantially across model families: Scaling law coefficients change with the pretraining data composition \citep[Section~3.3]{bi2024deepseek}, and data selection conclusions can flip when other hyperparameters are varied \citep{wang2025can}. 
When the observational analysis pools across families, it averages over this heterogeneity---the same cross-family replication that improves statistical precision can degrade external validity (Assumption~\ref{ass:external-validity:recipes}).
This is a fundamental tension for the observational approach:
Estimating recipe-specific effects requires parameters that describe the interaction of the treatment with other family-specific parameters, but fitting those parameters requires more data per family.

\textbf{Validity threats.} {\bred}~Due to confounding and selection bias, \emph{internal} validity is a key threat for the observational approach, requiring strong, untestable assumptions.
{\byellow}~What is more, depending on the research question, small sample sizes may threaten \emph{statistical} validity. Although aggregation across families may improve \emph{statistical} validity, this aggregation can come at the cost of \emph{external} validity when the treatment effect is heterogeneous.
{\bgreen}~Unless the approach is combined with proxies, it does not threaten construct validity.

%% file: neurips/sections/singlerun-v5.tex
\section{The Single-Run Approach}
\label{sec:single-run}

In this section, we continue our discussion of the validity threats of different research strategies with the single-run approach. The single-run approach emulates several experiments in a single training run, either by framing the run as involving a population of independently treatable units (Section \ref{sec:singlerun:within-run}), or by constructing theoretical control outcomes (Section  \ref{sec:singlerun:theoretical}). 
As we show below, the validity threats depend on the chosen strategy.

\subsection{Within-Run Control Outcomes}
\label{sec:singlerun:within-run}

\begin{figure}[t]
  \centering
  \includegraphics[width=\linewidth]{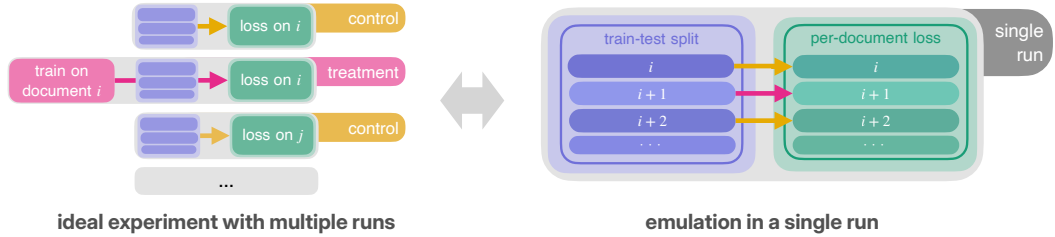}
  \caption{{\bf Analyzing a single training run.}
  Single-run analysis emulates multiple training runs (left) from one (right). Here, documents are regarded as independent ``units'', their assignment to the training set as the ``treatment'', and their respective loss as the ``outcome''.
  }
  \label{fig:onerun}
\end{figure}

For some research questions, we can frame one run as containing a population of qualitatively similar, \emph{independently treatable} units with \emph{separate} outcomes.
Consider data memorization, where the goal is to quantify the effect of training on a document on the model's behavior toward that document \citep{feldman2020neural,zhang2023counterfactual}.
Here, we can frame individual documents in the dataset as the ``units'', and their inclusion in the training set as the ``treatment''. 
Similarly, for tokenization bias, we aim to quantify the effect of representing a subword with its own token on the model's behavior toward that subword \citep{lesci2025causal}.
Here, the subwords are the ``units'', and inclusion in the vocabulary is the ``treatment''.
Using this framing, we can estimate \emph{average} causal effects across different units from a single run (Figure~\ref{fig:onerun}).

A naive estimator would simply compare average outcomes in the treatment and control group.
But this is only valid if the groups are \emph{exchangeable} %
(cf. Section~\ref{sec:observational}).
Although the training recipe does not differ between groups---resolving a central concern faced in Section~\ref{sec:observational}---exchangeability may be violated: Deduplication may differ between training and validation sets, and subwords are not randomly added to the vocabulary but systematically selected based on corpus frequency.
To address this, causal inference methods replace exchangeability with more plausible assumptions:
\citet{lesci2024causal} adopt a difference-in-differences (DiD) design to study memorization, which requires that both groups would have followed \emph{parallel outcome trends} in the absence of treatment.
For tokenization bias, \citet{lesci2025causal} exploit that the vocabulary cutoff is an arbitrary choice:
subwords just above and below this cutoff are \emph{quasi-randomly} assigned to treatment and control, enabling a regression discontinuity design (RDD).
 And \citet{elazar2022measuring} use a confounder adjustment to quantify the causal effect of pretraining token co-occurrence on the model's behavior.

Next to \emph{exchangeability}, a central threat for causal identification is \emph{no interference}.
No interference requires that treating one unit does not affect the outcomes of another (Assumption~\ref{ass:sutva}).
In the single-run setting, the assumption is strained:
all ``units'' share the same model parameters and training on data point $i$ changes the parameters also used to process data point $j$.
In contrast to exchangeability, interference received little discussion in the literature.

\textbf{Validity threats.} {\bred}~Interference across units and systematic differences between treatment and control group threaten \emph{internal} validity. Moreover, since the main trick is to \emph{average} across units we cannot quantify the effects for individual units, threatening \emph{external} validity (Assumption~\ref{ass:external-validity:units}). {\bgreen}~As long as within-run populations are large, the approach does not threaten \emph{statistical} validity.

\subsection{Theoretical Control Outcomes}
\label{sec:singlerun:theoretical}

To gain more fine-grained insight about the causal effects for individual units, \emph{theoretical controls} can be employed.
We discuss two settings:
One without experimental control over the model run, and one where we can inject data with known properties into training to obtain mathematical guarantees.

{\bf Observing vs.\ controlling one training run.}
In the first setting, theoretical controls are derived without any control over the training run.
One instance of this are Influence Functions, which are used to approximate the contribution of individual training examples to a particular prediction \citep{koh2017understanding,grosse2023studying,bae2024training,mlodozeniec2025distributional}.
To do so, they replace the effect of ``removing one data point'' with ``reducing the weight of one data point'', and employ strong statistical assumptions for its approximation.
While Influence Functions accurately capture the LOO retraining effect for simple models like linear or logistic regression \citep{hampel1974influence,giordano2019swiss, koh2017understanding}, the assumptions underlying the approximation fail for frontier models with highly non-convex loss landscapes and complex stochastic training dynamics \citep{schioppa2023theoretical,bae2022if,li2024influence}.

In the second setting, we control the training run, allowing us to \emph{inject} signals that can be used to derive a mathematically guaranteed control \citep{andrew2023one,pawelczyk2024machine,bordt2026train,sablayrolles2020radioactive,christ2024provably,bouaziz2024data}.
For example, in privacy auditing, \citet{steinke2023privacy} randomly assign ``canary'' data to the training or validation set.
The analytical control is derived from the binomial distribution governing these independent assignments. 
Applying a similar principle, \citet{pawelczyk2024machine} inject independent noise canaries (e.g., Gaussian canaries) into gradient updates, training data, or input embeddings \citep{pawelczyk2024machine,bordt2026train}. 
Because the expected dot product between a model's output and a Gaussian canary from the control group is analytically zero, the control baseline is mathematically guaranteed.
However, canary insertion may degrade model performance depending on signal strength---ranging from up to 5\% accuracy drop \citep{steinke2023privacy} to negligible impact when using Gaussian canaries \citep{pawelczyk2024machine}.

\textbf{Validity threats.} 
\bred~Interference (Assumption~\ref{ass:sutva}) remains a central threat for \emph{internal} validity: All units share the same model parameters, so training on canary $i$ may affect the outcome for canary $j$ \citep{keinan2025well}.
\byellow~Depending on the particular approach, \emph{statistical}, \emph{external}, and \emph{construct} validity may all be at risk:
As the case with influence functions, analytical approximations may rely on proxy treatments (\emph{construct} validity) and on strong statistical assumptions (\emph{statistical} validity). Injecting more ``canary'' data improves \emph{statistical} validity but modifies the training recipe and may reduce \emph{external} validity \emph{across models}.
\bgreen~However, analytical controls can be used to estimate effects for individual units, thereby maintaining \emph{external} validity \emph{across units}.

%% file: neurips/sections/validity-profiles.tex
\section{Validity Profiles for Foundation Model Research}
\label{sec:validity_profiles}

Sections~\ref{sec:proxys}--\ref{sec:single-run} analyzed the validity threats inherent to different foundation model research strategies.
Here, we synthesize these findings into characteristic {\bf validity profiles}.
For every research strategy, its validity profile states which types of validity are most at risk.
The different validity types are classified as being rather not affected (\bgreen), requiring careful consideration (\byellow), or key threat (\bred). 
Importantly, a green indicator does not guarantee a research design that follows the strategy is automatically valid---it simply signifies that the strategy does not introduce structural vulnerabilities of that type.

Table~\ref{tab:validity_indicators} summarizes the characteristic validity profiles of each research strategy.
\emph{Proxy approaches} preserve statistical and internal validity by design and admit a single, targeted threat---external validity for proxy models, and construct validity for proxy treatments and outcomes.
\emph{The observational approach} retains evidence at target scale but introduces an untestable internal validity threat (confounding), alongside additional statistical and external validity concerns.
\emph{Single-run designs} share an internal validity threat from interference between treated units, trading external validity across units against other validity types.
Importantly, no row in the matrix is uniformly green. 
{\it Since all approaches approximate the ideal experiment with less compute, they all rely on assumptions whose violation invalidates conclusions.}

In practice, the choice of research strategy is often largely determined by the specific research question being asked.
Therefore, the profiles depicted in Table~\ref{tab:validity_indicators} serve a dual purpose.
For authors, they provide a vocabulary to make hidden assumptions explicit and defend the necessary trade-offs their design entails.
For reviewers, the profiles provide a structured basis to assess whether a paper's conclusions are supported by its underlying methodology.
This becomes especially important when several strategies are combined within one study, for example when an observational analysis uses a proxy treatment to impute missing meta-data. In this case, the validity threats accumulate. We argue that 
researchers must therefore carefully align their claims with the active threats and, where feasible, pursue multiple strategies in parallel to strengthen the conclusions \citep{staley2004robust,claveau2019variety}.

\begin{table}[t]
\centering
\footnotesize
\caption{{\bf Validity profiles of different research strategies for foundation models.}
Each cell encodes how the corresponding validity type (column, Section~\ref{sec:the_framework}) fares under the research strategy (row, Sections~\ref{sec:proxys}--\ref{sec:single-run}).
Each cell summarizes how the strategy affects the corresponding validity type: \bgreen~rather not affected; \byellow~careful consideration required; \bred~key threat.
}
\label{tab:validity_indicators}
\renewcommand{\arraystretch}{1.2}
\begin{tabular*}{\linewidth}{@{}l@{\extracolsep{\fill}}c c c c@{}}
\toprule
\textbf{Strategy} & \textbf{Statistical} & \textbf{Internal} & \textbf{External} & \textbf{Construct} \\
\midrule

\multicolumn{5}{@{}l}{\textbf{Proxy approach} \textit{(Section~\ref{sec:proxys})}} \\
\quad Proxy models            & \bgreen  & \bgreen   & \bred & \bgreen   \\
\quad Proxy treatments and outcomes      & \bgreen  & \bgreen   & \bgreen   & \bred \\
\addlinespace %
\textbf{Observational approach} \textit{(Section~\ref{sec:observational})}    & \byellow   & \bred & \byellow    & \bgreen   \\
\addlinespace
\multicolumn{5}{@{}l}{\textbf{Single-run approach} \textit{(Section~\ref{sec:single-run})}} \\
\quad Within-run control outcomes  & \bgreen  & \bred & \bred & \bgreen   \\
\quad Theoretical control outcomes    & \byellow   & \bred & \byellow    & \byellow \\
\bottomrule
\end{tabular*}
\end{table}

%% file: neurips/sections/related_work.tex
\section{Related Work}
\label{sec:related-work}

Traditionally, machine learning relied on the ability to conduct (repeated) controlled experiments at the target scale. Within this paradigm, discussions of validity have usually concentrated on the \emph{outcome end} of the pipeline. In particular, a large number of works discuss the (construct) validity of benchmarks \citep{jacobs2021measurement,raji2021ai,bowman2021what,dehghani2021benchmark,wallach2024evaluating,freiesleben2025benchmarking,bean2025measuring}, the (statistical) validity of model comparisons \citep{bouthillier2021accounting,miller2024error,madaan2024quantifying,herrmann2024position}, and more generally the evaluation of machine learning systems \citep{liao2021learning,pineau2021improving,reuel2024betterbench,hardt2025emerging}. The validity challenges discussed in these works are acutely relevant for foundation model research and often persist even when we are able to perform the ideal experiment. As such, they are mostly orthogonal to the challenges that we discuss in this work. 

An increasing number of works discuss validity challenges in foundation model research. For example, \citet{koh2025predicting} discuss the external validity of proxy models for reasoning benchmarks, and \citet{zhang2025position} argue that membership inference is not a valid method to detect training data inclusion. Many other examples are discussed in Sections \ref{sec:proxys}-\ref{sec:single-run}. Prior works usually focus on individual research questions and designs. In contrast, this work highlights the similarities across {\it different} research designs and offers a framework that can be used to study validity beyond individual research questions.

%% file: neurips/sections/supplement.tex
\newpage
\appendix

\section{Glossary}
\label{app:glossary}

Table~\ref{tab:glossary} summarizes the central notation used throughout the paper; this section gives intuitive definitions for the most important terms.

\begin{table}[htbp]
    \centering
    \caption{Overview of the notation used in the paper.}
    \label{tab:glossary}
    \begin{tabular}{lll}
        \toprule
        Notation & Terminology & Meaning \\
        \midrule
        $\recipe$ & training recipe & all controllable parts of training \\
        & & (architecture, optimizer, training data, $\ldots$) \\
        $\outcome$ & outcome & any property we care about \\
        & & (training loss, benchmark, downstream performance, $\ldots$) \\
        $\treatment$ & treatment & a change to the recipe \\
        $\effect$ & treatment effect & change in outcome induced by the treatment \\
        $\effect(\treatment, \outcome; \recipe)$ & recipe-specific treatment effect & see Definition~\ref{def:ate-psi} \\
        $\recipe'$ & proxy recipe & e.g.\ with smaller model size \\
        $\treatment'$ & proxy treatment & e.g.\ fine-tuning instead of pre-training \\
        $\outcome'$ & proxy outcome & e.g.\ a cheap benchmark approximation \\
        $\beta$ & statistical estimand & a parameter of the data distribution \\
        \bottomrule
    \end{tabular}
\end{table}

\textbf{Causal effect ($\effect$).} The change in the outcome induced by the treatment, holding the rest of the recipe fixed. The causal effect is what we ultimately care about---classically obtained by means of a controlled experiment, and increasingly by means of an approximation strategy (Sections~\ref{sec:proxys}--\ref{sec:single-run}). Eventually we are interested in the recipe-specific average treatment effect (Definition~\ref{def:ate-psi}). When we want to emphasize the inferential role of $\effect$ in contrast to the data-distributional $\beta$ below, we call it the \emph{causal estimand}.

\textbf{Causal identification.} The process of translating a \emph{causal estimand} (the treatment effect $\effect$) into a statistical estimand (in the best case, $\beta$). To do so, causal identification must rely on assumptions about the underlying causal structure. We explain one example in Appendix~\ref{app:internal}.

\textbf{Confounding.} A non-causal statistical dependence between treatment and outcome induced by a common cause of both. Confounding is one mechanism through which exchangeability (Assumption~\ref{ass:exchangeability}) can fail; we discuss it in Section~\ref{sec:observational} and illustrate it in Figure~\ref{fig:observational:graph}.

\textbf{Ideal experiment.} The hypothetical experiment that identifies the causal effect of interest by training many models with and without the treatment and comparing average outcomes. At frontier scale this is infeasible because we cannot afford to train many models; Sections~\ref{sec:proxys}--\ref{sec:single-run} discuss the approximation strategies that researchers use instead.

\textbf{Proxy model ($\recipe'$).} A model trained according to a scaling recipe $\recipe'$ that follows the original training recipe $\recipe$ but uses substantially less compute (Section~\ref{sec:proxy:models}).

\textbf{Proxy treatment ($\treatment'$) and proxy outcome ($\outcome'$).} Computationally cheaper substitutes for the target treatment or outcome (Section~\ref{sec:proxy:treatments}). A proxy treatment may be qualitatively different from the original (e.g.\ fine-tuning in place of pre-training); a proxy outcome may be an approximation of the target (e.g.\ a benchmark subset) or altogether different (e.g.\ pretraining loss in place of post-training benchmark scores).

\textbf{Scaling recipe.} A specification of how the training recipe changes as it is scaled up or down: how the learning rate, architecture, and other hyperparameters are adjusted as the model and training-data sizes change. The GPT-3 paper provides a canonical example \citep[Table~2.1]{brown2020language}; many scaling-law papers use slightly customized recipes \citep{kaplan2020scaling,hoffmann2022training}.

\textbf{Selection bias.} A non-causal statistical dependence between treatment and outcome induced by the process by which observations enter the sample, e.g.\ the decision to publish a model. In line with recent causal inference literature \citep{hernan2016using}, we use ``selection bias'' in the narrow sense where conditioning on the selection introduces a noncausal association between treatment and outcome; see Appendix~\ref{app:internal} for an extended discussion and Figure~\ref{fig:observational:graph} for the corresponding causal graph.

\textbf{Statistical estimand ($\beta$).} A parameter of the observable data distribution---e.g.\ a conditional expectation, a regression coefficient, a difference of group means.

\textbf{Training recipe ($\recipe$).} The collection of all controllable parts of model training---architecture, optimizer, learning-rate schedule, training data, random seeds, and so on. The training recipe contains everything required to replicate an \emph{individual} training run.

\textbf{Treatment ($\treatment$) and outcome ($\outcome$).} A treatment is the modification of interest---typically a change to the training recipe, e.g.\ adding math data to the pre-training mix or increasing the learning rate. The outcome is the property we wish to influence---e.g.\ benchmark performance, training loss, or a privacy metric. We adopt the potential-outcomes framework \citep{rubin1974estimating} to formalize the outcomes under alternative treatments (but equivalent formulations using Pearl's do-operator are possible \citep{pearl2009causality}).

\textbf{Validity.}
Conclusions can become invalid for various reasons.
Based on a framework from the causal inference literature in the social sciences \citep{shadish2001validity} we distinguish four types of validity: \emph{statistical}, \emph{internal}, \emph{external}, and \emph{construct} (Section~\ref{sec:the_framework}).
A more formal introduction of the validity types is provided in Appendix~\ref{app:formal}.

\textbf{Validity-profile bullets ({\bgreen}, {\byellow}, {\bred}).} In Sections~\ref{sec:proxys}--\ref{sec:single-run} and Table~\ref{tab:validity_indicators} we mark each (strategy, validity type) pair with a coloured bullet: {\bgreen} indicates that the strategy per se does not threaten the validity type; {\byellow} that the strategy raises threats requiring careful consideration in practice; and {\bred} that the strategy poses a key threat to the validity type.

\section{Formalizing the Assumptions Made by Different Research Strategies}
\label{app:formal}

As follows, we provide a more formal treatment of key assumptions made by the different research strategies. We structure the section according to the four validity types (Appendix~\ref{app:statistical}-\ref{app:construct}).

\paragraph{A side note on scope.}
Throughout this paper, our concern is whether the \emph{approximation strategy} introduces \emph{new} validity threats. Thus, assumptions that are made by some original experiment that is approximated with a particular research strategy---e.g.\ the construct validity of the chosen benchmark---are orthogonal to our analysis. 
We do not discuss them here.

\subsection{Statistical Validity}
\label{app:statistical}

Every research strategy eventually approximates some population parameter $\beta$ from a finite sample. The form of $\beta$ depends on the strategy: in a controlled experiment with one treatment and one control run, the estimate $\hat{\beta}$ could be the different between the two outcomes, 
and the estimand the \emph{expected} difference between the two outcomes.
In an observational analysis, 
the estimate may be a regression coefficient $\hat{\beta}$,
and the estimand the 'true' population parameter.
In all cases, statistical validity asks whether the estimate $\hat\beta$ from the finite sample reliably reflects the underlying population parameter,
\begin{equation*}
  \hat\beta \approx \beta.
\end{equation*}
This approximation can rely on strong assumptions; for example a linear model assumes a linear relationship between treatment and outcome. Since this is standard knowledge in the field, we do not discuss these assumptions in more detail here.

\subsection{Internal Validity}
\label{app:internal}

Suppose we have estimated $\beta$ accurately---for example, we may have found a strong association between using a new architecture and some benchmark score. The question for internal validity is whether this association reflects a genuine causal effect, or, slightly more formally, whether

\begin{equation*}
  \beta = \effect.
\end{equation*}

In a well-designed controlled experiment, $\beta = \effect$ typically holds \emph{by design}. Ablations, hyperparameter sweeps, and grid searches randomize over the runs that produce treatment and control groups, so systematic differences between the groups vanish in expectation, and the difference of group means identifies the causal effect.
\footnote{As we make formal below, this is because the identifying assumptions---exchangeability, positivity, and SUTVA---are usually satisfied by construction in such experiments.}

\subsubsection{Internal Validity in Section~\ref{sec:observational}}
\label{app:internal:observational}

In observational data, the equality $\beta = \effect$ must be \emph{argued for} using a set of structural assumptions about the data-generating process, a step known as \emph{causal identification} \citep{rubin1974estimating,pearl2009causality}. A rich literature provides identification tools---the back-door and front-door criteria, instrumental variables, difference-in-differences, regression discontinuity, do-calculus---each requiring different assumptions and applicable in different settings. Most designs in the observational approach (Section~\ref{sec:observational}) take the simplest stance and assume that the parameters learned by a treatment-outcome association model are themselves causal: that the pairwise association between $\treatment$ and $\outcome$ already coincides with the causal effect.
Therefore, three assumptions must hold: exchangeability, positivity, and SUTVA.
Since these assumptions did not receive attention in the primary literature, we introduce them more formally below.

\begin{figure}[htbp]
    \centering
    \includegraphics[width=0.9\textwidth]{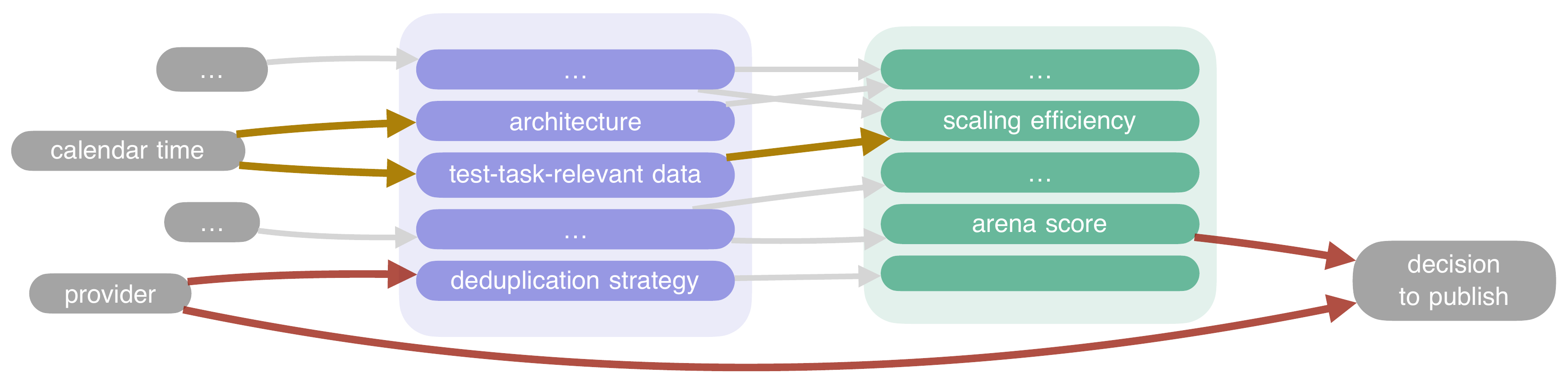}
    \caption{{\bf Confounding and selection bias in the observational approach.}
    In observational data, \textcolor{recipecolor}{recipe components} and \textcolor{outcomecolor}{outcomes} can be statistically dependent via non-causal paths.
    \emph{Confounders} are variables that cause both the treatment and outcome thereby inducing a non-causal dependence between them. In our example, calendar time confounds the relationship between architectural choices and scaling efficiency via training on task-relevant data (\textcolor{brown}{brown path}).
    \emph{Selection bias} arises from how observations enter the sample. The decision to publish acts as a \emph{collider}, which opens non-causal paths between recipe components and outcomes. In our example, the training recipe choices of certain providers appear more favorable in part because these providers can pick and publish the best score out of several (noisy and private) benchmark runs (\textcolor{BrickRed}{dark red path}).
    }
    \label{fig:observational:graph}
\end{figure}

\paragraph{Exchangeability.}
Exchangeability (Definition~\ref{ass:exchangeability}) requires that the units that receive a particular treatment do not differ systematically from those that do not, so the difference of group means is causally interpretable.
The two canonical examples by which exchangeability fails are confounding (a common cause of $\treatment$ and $\outcome$) and selection bias (conditioning on a collider); both are illustrated in Figure~\ref{fig:observational:graph}.

We illustrated the case of confounding in detail in Section~\ref{sec:observational}.
As follows we discuss \emph{selection bias}.
While confounding concerns the data-generating process itself, selection bias arises from the process by which observations are \emph{included} in the sample.\footnote{We use ``selection bias'' in the narrow Pearl/Hern\'an sense of conditioning on a collider \citep{hernan2016using}; some older causal-inference literature uses the term as an umbrella for any violation of exchangeability, including confounding.} For example, \citet{singh2025leaderboard} show that some labs evaluate many model variants on private benchmark data before publishing the best score. Conditioning on the implicit ``decision to publish'' opens a non-causal path between provider and benchmark score: a high observed score is more likely to come from a provider that could privately test several variants beforehand. 

In Definition~\ref{ass:exchangeability} we formally introduce exchangeability. In the simplest case there are two potential outcomes, one for the treatment $\treatment$ and one for the control $\control$;
this is the case that we discuss in the main text.
Below we use more general definitions that allow a range of possible treatments $\treatment \in \mathcal{A}$.

\begin{assumption}[Exchangeability]
\label{ass:exchangeability}
The potential outcome $\textcolor{outcomecolor}{Y}(\treatment)$ under treatment value $\treatment$ is independent of the observed treatment $A$:
\[
\textcolor{outcomecolor}{Y}(\treatment) \idp A \quad \text{for all } \treatment \in \mathcal{A}.
\]
\end{assumption}

\paragraph{Positivity.}
Even if exchangeability holds, the causal effect of a particular treatment value cannot be estimated unless that value is observed in the sample. In the simple setting with one treatment $\treatment$ and one control $\control$, positivity is typically satisfied.
In this setting, the assumption can also easily be verified based on a sample.

\begin{assumption}[Positivity]
\label{ass:positivity}
Every treatment value of interest must occur with positive probability in the data:
\[
P(A = \treatment) > 0 \quad \text{for all } \treatment \in \mathcal{A}.
\]
\end{assumption}

\paragraph{SUTVA.}
The third assumption is the \emph{stable unit treatment value} assumption, which has two parts: \emph{no interference} between units, and \emph{consistency} between observed and potential outcomes.

\begin{assumption}[SUTVA]
\label{ass:sutva}
Let $\treatment_i$ denote the treatment received by unit $i$ and $\outcome_i$ the corresponding observed outcome. SUTVA comprises:
\begin{enumerate}
  \item \emph{No interference:} $\textcolor{outcomecolor}{Y}_i(\treatment_1, \ldots, \treatment_n) = \textcolor{outcomecolor}{Y}_i(\treatment_i)$ for all $(\treatment_1, \ldots, \treatment_n) \in \mathcal{A}^n$, i.e.\ the potential outcome of unit $i$ depends only on its own treatment.
  \item \emph{Consistency:} $A_i = \treatment \implies Y_i = Y_i(\treatment)$, i.e.\ the observed outcome equals the potential outcome under the treatment actually received.
\end{enumerate}
\end{assumption}

In the observational approach, a violation of no interference means that the treatment received by one training run influences the potential outcomes of other runs. For instance, if the fact that one model was trained with the muon optimizer changes the potential outcomes (both with and without muon) for another model, no interference is violated. In the single-run setting the units are not the training runs but the documents. Thus no interference has a slightly different interpretation in this context (see below).

\subsubsection{Assumptions Made by Research Strategies in Section \ref{sec:single-run}}
\label{app:internal:single-run}

The single-run designs in Section \ref{sec:single-run} relax the unconditional exchangeability of Assumption~\ref{ass:exchangeability} and replace it with an alternative identifying assumption that is more credible in the single-run setting. \citet{lesci2024causal} adopt a difference-in-differences design, which replaces exchangeability with a \emph{parallel outcome trends} assumption: in the absence of treatment, the expected change in outcome over training time is the same in the treated and control groups. \citet{lesci2025causal} use a regression discontinuity design, which assumes that potential outcomes vary continuously across the cutoff that determines treatment assignment, so units just above and below the cutoff are exchangeable in the limit. \citet{elazar2022measuring} instead invoke \emph{conditional exchangeability}: potential outcomes are independent of treatment after conditioning on a sufficient set of observed covariates. We refer to the original papers for the formal definitions.

It is important to note that relaxing exchangeability typically tightens the other assumptions required for identification. For example, conditional exchangeability requires \emph{conditional positivity}: every treatment value must occur with positive probability \emph{within every stratum} of the conditioning covariates---a substantially stronger requirement than unconditional positivity (Assumption~\ref{ass:positivity}).

For Section~\ref{sec:single-run}, \emph{no interference} (Assumption~\ref{ass:sutva}) is particularly relevant. 
In the single run setting, the ``treatments'' for the different ``units'', for example training on a particular document or not, all affect the same training (Sections~\ref{sec:singlerun:within-run} and \ref{sec:singlerun:theoretical}): training on data point $i$ updates the parameters that are subsequently used to process data point $j$, so the potential outcome for $j$ depends on $i$'s treatment.
As such, the assumption is strained.

\subsection{External Validity}
\label{app:external}

External validity concerns the generalization of a causal effect: even if the effect $\effect$ is correctly identified for the recipe under study, does it apply to the recipe (or population, or unit) we ultimately care about? We distinguish several flavors of external-validity claim that recur throughout the paper: external validity \emph{across recipes}, in particular \emph{across scale}, and external validity \emph{across units}.

\subsubsection{Assumptions Made by Research Strategies in Section \ref{sec:proxy:models}}
\label{app:external:proxy-models}

\paragraph{External validity across families and scale.}
The recipe-specific treatment effect (Definition~\ref{def:ate-psi}) depends on the recipe $\recipe$ at which it is evaluated. 
When the treatment effect is evaluated for a proxy $\recipe'$ it is unclear whether the causal effect also applies for our target recipe $\recipe$.
We distinguish a couple of special cases.
External validity \emph{across scale} considers the case where the proxy $\recipe'$ is a scaled-down variant of the original $\recipe$.
 External validity \emph{across families} is the special case in which $\recipe$ and $\recipe'$ belong to different model families.
In both cases, we want the treatment effect for the proxy and original recipe to be the same.

\begin{assumption}[\bf External validity of a proxy recipe]
\label{ass:external-validity:proxy-recipe}
Let $\recipe$ be the target training recipe and $\recipe'$ a proxy training recipe; let $\treatment$ be the treatment of interest and $\outcome$ the outcome. The proxy recipe is \emph{externally valid} with respect to $(\treatment, \outcome, \recipe)$ iff
\begin{equation*}
  \effect(\treatment, \outcome; \recipe) = \effect(\treatment, \outcome; \recipe').
\end{equation*}
\end{assumption}

\subsubsection{Assumptions Made by Research Strategies in Section \ref{sec:observational}}
\label{app:external:observational}

\paragraph{External validity across recipes.}
The observational approach (Section \ref{sec:observational}) often pools data across many model families to obtain enough statistical power. Even if the resulting estimate is causally identified, it does not target the recipe-specific effect of Definition~\ref{def:ate-psi}, but rather an \emph{average} effect over the distribution of recipes from which the sample is drawn. External validity then concerns whether this average matches the recipe-specific effect for the target recipe.

\begin{assumption}[\bf External validity of an average effect across recipes]
\label{ass:external-validity:recipes}
Let $\Pi$ be a distribution over training recipes (e.g.\ the empirical distribution over publicly documented model families) and let
\[
  \bar\effect(\treatment, \outcome; \Pi) := \mathbb{E}_{\recipe \sim \Pi}\bigl[\effect(\treatment, \outcome; \recipe)\bigr]
\]
be the average treatment effect across recipes drawn from $\Pi$. The average effect $\bar\effect(\treatment, \outcome; \Pi)$ is \emph{externally valid} with respect to the target recipe $\recipe$ iff
\begin{equation*}
  \effect(\treatment, \outcome; \recipe) = \bar\effect(\treatment, \outcome; \Pi).
\end{equation*}
\end{assumption}

The average and the recipe-specific effect coincide when the treatment effect is homogeneous across recipes in the support of $\Pi$, and may diverge when the effect interacts with other recipe components (cf.\ the heterogeneity discussion in Section~\ref{sec:proxy:models}).

\subsubsection{Assumptions Made by Research Strategies in Section \ref{sec:single-run}}
\label{app:external:single-run}

\paragraph{External validity across units.}
We face a structurally similar problem in Section~\ref{sec:singlerun:within-run}.
But in the single run setting, the population are not the different training runs/recipes $\recipe^{(i)}$, but the \emph{components} of the recipe $\recipe_i$, for example the training/validation set assignment of a document $i$ in the dataset.
In the context of the within-run approach, the question is whether averages across those within-run units (documents) reflect the effect for a particular within-run unit.

\begin{assumption}[\bf External validity of an average effect across within-run units]
\label{ass:external-validity:units}
Let $\effect_i$ denote the unit-specific treatment effect for unit $i$ and $\bar\effect := \frac{1}{n}\sum_{i=1}^n \effect_i$ the average across units.
The average effect $\bar\effect$ is externally valid with respect to a target unit $i$ iff
\begin{equation*}
  \effect_i = \bar\effect.
\end{equation*}
\end{assumption}

As with external validity across recipes, the average and the unit-specific effect coincide when effects are homogeneous across units and may diverge under heterogeneity. In Section \ref{sec:singlerun:theoretical} we also discuss external validity across models (Assumption \ref{ass:external-validity:proxy-recipe}).

\subsection{Construct Validity}
\label{app:construct}

In general, the line between external and construct validity can be blurry \citep[Ch.~3]{shadish2001validity};
Both concern some form of generalization after all.
But while external validity concerns generalization across populations, construct validity is about generalization across operationalizations of constructs.

We make a clear distinction: In our discussion \emph{external validity} concerns whether a causal effect transfers across recipes or recipe components \emph{for the same causal question}, while \emph{construct validity} concerns whether the causal question itself is preserved when the treatment or outcome are replaced by a proxy. As follows, we formalize construct validity for proxy treatments and proxy outcomes.

\subsubsection{Assumptions Made by Research Strategies in Section \ref{sec:proxy:treatments}}
\label{app:construct:proxy-treatments}

\begin{assumption}[\bf Construct validity of a proxy treatment]
\label{ass:construct-validity:proxy-treatment}
Let $\treatment$ be the target treatment and $\treatment'$ a proxy treatment; let $\outcome$ be the outcome and $\recipe$ the recipe. The proxy treatment is \emph{construct-valid} with respect to $(\treatment, \outcome, \recipe)$ iff
\begin{equation*}
  \effect(\treatment, \outcome; \recipe) = \effect(\treatment', \outcome; \recipe).
\end{equation*}
\end{assumption}

\begin{assumption}[\bf Construct validity of a proxy outcome]
\label{ass:construct-validity:proxy-outcome}
Let $\outcome$ be the target outcome and $\outcome'$ a proxy outcome; let $\treatment$ be the treatment and $\recipe$ the recipe. The proxy outcome is \emph{construct-valid} with respect to $(\treatment, \outcome, \recipe)$ iff
\begin{equation*}
  \effect(\treatment, \outcome; \recipe) = \effect(\treatment, \outcome'; \recipe).
\end{equation*}
\end{assumption}

Assumptions~\ref{ass:construct-validity:proxy-treatment}--\ref{ass:construct-validity:proxy-outcome} take the original $\treatment$ and $\outcome$ as fixed; the prior question of whether $\treatment$ and $\outcome$ themselves operationalize the underlying constructs (e.g.\ whether ``MMLU score'' faithfully operationalizes ``general reasoning ability'') is---per the section opener---orthogonal to the threats introduced by the approximation. We refer the reader to recent work on benchmark validity in machine learning for treatments of that question \citep{bean2025measuring,freiesleben2025benchmarking,wallach2024evaluating}.